\documentclass[10pt,twocolumn,letterpaper]{article}

\usepackage[table,xcdraw]{xcolor}
\usepackage{multirow}
\newsavebox{\mybox}

\usepackage{cvpr}                
\definecolor{cvprblue}{rgb}{0.21,0.49,0.74}
\usepackage[pagebackref,breaklinks,colorlinks,allcolors=cvprblue]{hyperref}
\usepackage[accsupp]{axessibility}

\def\paperID{} 
\def\confName{CVPR}
\def\confYear{2026}

\title{MDS-VQA: Model-Informed Data Selection for Video Quality Assessment}

\author{
    Jian Zou$^{1}$, \
    Xiaoyu Xu$^{1}$, \
    Zhihua Wang$^{1}$, \
    Yilin Wang$^{2}$, \
    Balu Adsumilli$^{2}$, 
	and Kede Ma$^{1}$\thanks{Corresponding author.}\footnotemark[1] \\
	$^1$City University of Hong Kong \ $^2$Google Inc. \\
	{\tt\small jian.zou@my.cityu.edu.hk,} 
    {\tt\small \{yilin, badsumilli\}@google.com,} \\
    {\tt\small \{xiaoyxu, zhihua.wang, kede.ma\}@cityu.edu.hk} \\
    {\tt\normalsize \href{https://github.com/Multimedia-Analytics-Laboratory/MDS-VQA}{\textcolor{magenta}{https://github.com/Multimedia-Analytics-Laboratory/MDS-VQA}}}
}

\begin{document}
\maketitle
\begin{abstract}

Learning-based video quality assessment (VQA) has advanced rapidly, yet progress is increasingly constrained by a disconnect between model design and dataset curation. Model-centric approaches often iterate on fixed benchmarks, while data-centric efforts collect new human labels without systematically targeting the weaknesses of existing VQA models. Here, we describe MDS-VQA, a model-informed data selection mechanism for curating unlabeled videos that are both difficult for the base VQA model and diverse in content. Difficulty is estimated by a failure predictor trained with a ranking objective, and diversity is measured using deep semantic video features, with a greedy procedure balancing the two under a constrained labeling budget. Experiments across multiple VQA datasets and models demonstrate that MDS-VQA identifies diverse, challenging samples that are particularly informative for active fine-tuning. With only a $5\%$ selected subset per target domain, the fine-tuned model improves mean SRCC from $0.651$ to $0.722$ and achieves the top gMAD rank, indicating strong adaptation and generalization.

\end{abstract}    
\section{Introduction}
\label{sec:intro}

\begin{figure}[!t]
	\centering
	\includegraphics[width=\linewidth]{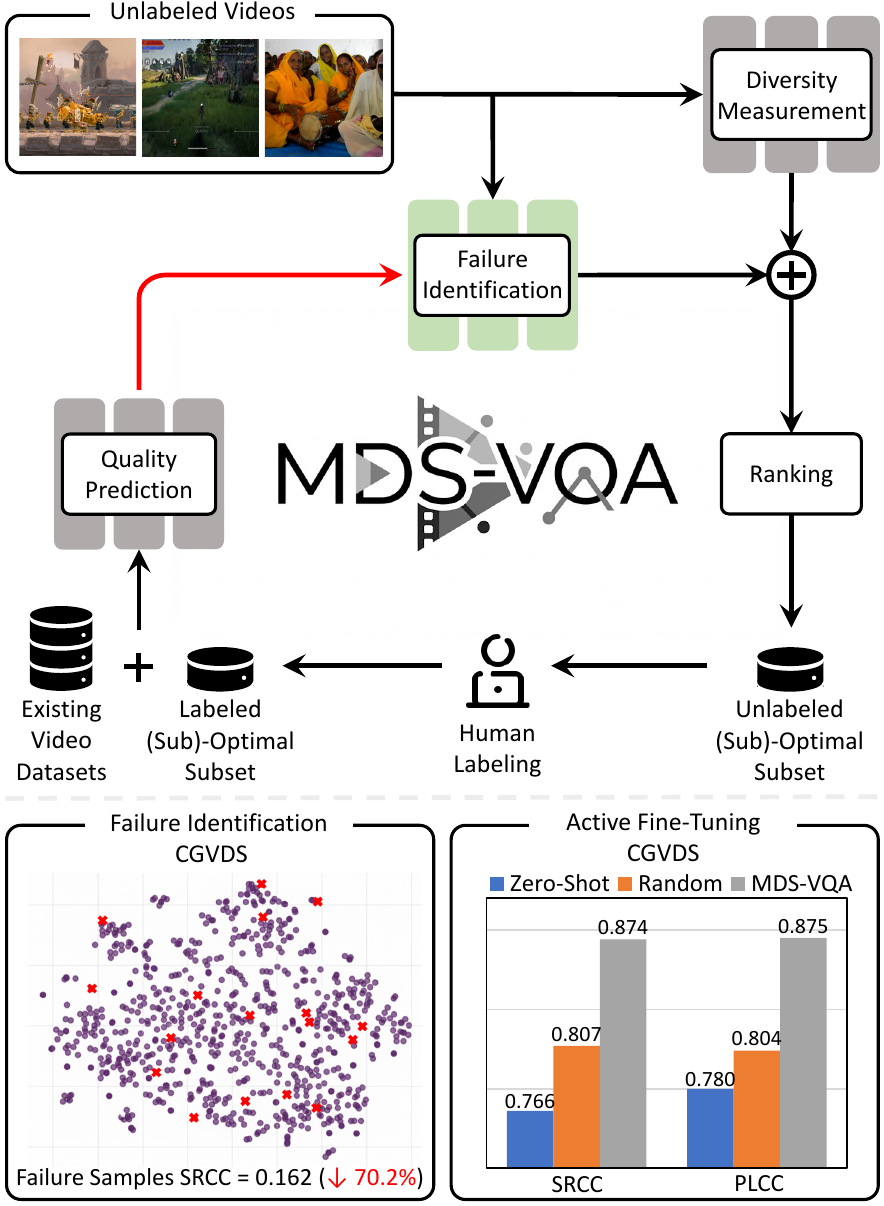}
	\caption{System diagram of MDS-VQA. We predict failure (\ie, difficulty) on unlabeled target videos, combine difficulty and content diversity to select a small subset for human labeling, and actively fine-tune the VQA model on existing labeled data and the selected subsets. Bottom: on CGVDS~\cite{zadtootaghaj2020quality}, this $5\%$ subset (marked by red crosses) yields strong failure identification and improved fine-tuning performance.
    }
	\label{fig:teaser}
\end{figure}

Video quality assessment (VQA) aims to predict perceptual video quality in a manner consistent with human judgments. Over the past two decades, VQA models have evolved from hand-crafted feature engineering tailored to specific distortion families~\cite{wang2000blind, tong2004blur} to deep neural networks that learn perceptual cues directly from data~\cite{korhonen2020blind, wu2022fast, wen2024modular}, and more recently to large-scale pretrained models that leverage rich visual prior knowledge to achieve more human-aligned quality reasoning~\cite{wu2025visualquality}.

Despite this architectural progress, the field is increasingly constrained by a structural mismatch between \textit{model-centric} innovation and \textit{data-centric} curation.  
Model-centric VQA continuously refines architectures~\cite{wang2021rich, ying2021patch}, loss functions~\cite{liu2018end, chen2021unsupervised}, and training recipes~\cite{li2022blindly, li2021unified, wu2022fast}, but typically does so on a small set of heavily reused benchmarks, creating persistent pressure to overfit dataset-specific idiosyncrasies. In parallel, data-centric VQA expends substantial resources on new subjective studies~\cite{wang2019youtube, zadtootaghaj2020quality, shang2021assessment, wang2024youtube, wang2025aigv} to collect mean opinion scores (MOSs), yet these efforts often proceed without systematically targeting the failure modes of current top-performing VQA models~\cite{cao2024image, wang2021active, wang2021troubleshooting}. 

A direct consequence of this inefficient loop is the \textit{easy dataset} problem in VQA~\cite{sun2024analysis}. When datasets are dominated by content with easily identifiable distortions, even simple baselines (without spatiotemporal analysis and aggregation) can perform competitively, obscuring the limitations of advanced model architectures and reducing the marginal value of collecting more labels of the ``same kind.'' As a result, new data collection---while valuable---does not always illuminate the true ``blind spots'' of contemporary VQA methods, and thus does not reliably drive improvements in cross-domain generalization.

In this work, we introduce MDS-VQA, a model-informed data selection mechanism for VQA that explicitly closes the feedback loop from models back to data (see Fig.~\ref{fig:teaser}). The key idea is to make dataset curation \textit{model-aware}: rather than sampling videos for annotation passively or purely based on representativeness, we prioritize videos that are 1) \textit{difficult} for the base model and 2) \textit{diverse} in content. Concretely, we augment a base VQA model with an auxiliary failure predictor that estimates the difficulty of each unlabeled video (\ie, how likely it is to expose the model's errors). We then combine these difficulty estimates with a content diversity measure to select a budgeted subset for human labeling. Active fine-tuning on the resulting ``hard-and-diverse'' subset completes the loop: model weaknesses guide data acquisition, and the acquired data directly improves the model.  

Extensive experiments across multiple VQA datasets and models confirm that MDS-VQA is consistently more effective than competing data selection strategies at both 1) identifying failure samples and 2) improving quality prediction performance after active fine-tuning. In particular, using only a $5\%$ selected subset per target domain, the fine-tuned model improves mean Spearman’s rank correlation coefficient (SRCC) from $0.651$ to $0.722$, and achieves the top rank in a group maximum differentiation (gMAD) competition~\cite{ma2018group}, supporting robust generalization beyond average-case correlation metrics.

\section{Related Work}
\label{sec:related}

In this section, we briefly summarize prior work on VQA, covering both model-centric and data-centric advances.

\begin{figure*}[!htb]
	\centering
	\includegraphics[width=\textwidth]{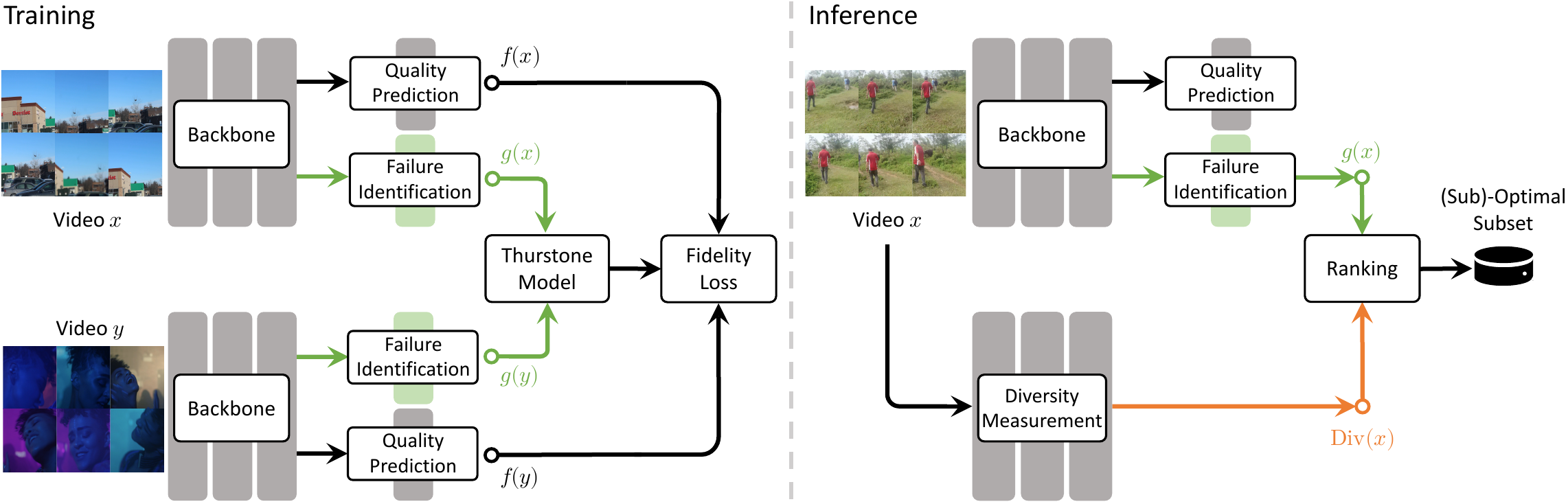}
	\caption{Training and inference of MDS-VQA. During training, we freeze the base quality model $f(\cdot)$ and optimize an auxiliary failure predictor $g(\cdot)$ by minimizing a fidelity loss under a Thurstone model~\cite{thurstone1927law}. During inference, we rank unlabeled videos by combining predicted difficulty scores with a content diversity measure to select a (sub-)optimal subset for human labeling and active fine-tuning.
    }
	\label{fig:overview}
\end{figure*}

\subsection{Model-Centric VQA}

The evolution of model-centric VQA reflects a continuous search for richer representations of perceptual quality. 
Early knowledge-driven methods~\cite{mittal2012no} rely on hand-crafted features, often grounded in natural scene statistics, to capture statistical regularities in visual signals, but they struggle with the complex, mixed distortions typical of real-world videos. 
With deep learning, VQA moves from manual feature engineering to learned representations. Early deep VQA models~\cite{le2006convolutional} adopt 2D convolutional neural network (CNN) backbones, processing frames independently and aggregating temporal information afterward. 
To more directly encode spatiotemporal structure, 3D CNNs~\cite{tran2015learning} are leveraged to directly learn from video volumes using 3D kernels~\cite{wang2021rich}, yet their limited receptive fields may hinder modeling long-range temporal dependencies. Transformer-based approaches~\cite{liu2022video} address this limitation by leveraging attention computation to capture global dependencies over long sequences, as exemplified by FAST-VQA~\cite{wu2022fast}.

More recently, vision-language models (VLMs) have been fine-tuned for VQA using quality scores, distributions, and/or descriptions, achieving improved performance. Subsequent work further applies reinforcement learning to optimize VLMs for VQA~\cite{zhang2025vq,wu2025visualquality}, by aligning model outputs with MOSs to improve quality reasoning and scoring. While many reinforcement learning-based methods still retain a regression-style objective for quality prediction, VisualQuality-R1~\cite{wu2025visualquality} is a notable example that instead adopts a reinforcement-learning-to-rank formulation, which naturally handles varying MOS scales across datasets and supports efficient adaptation under limited supervision. For these advantages, we adopt VisualQuality-R1 to implement our base VQA model in the main experiments to push performance under a strong, state-of-the-art VLM setting.

\subsection{Data-Centric VQA}

Data-centric VQA is mainly driven by collecting human perceptual judgments of videos, mostly MOSs or difference MOSs. Such annotations require time-consuming subjective studies~\cite{ghadiyaram2017capture, seshadrinathan2010study}, making large-scale, reliable datasets costly in both time and human effort.
Early VQA datasets~\cite{seshadrinathan2010study} are typically built from professionally generated content by applying synthetic distortions (\eg, noise, blur, and compression artifacts) to a small set of pristine source videos. While crucial for early development, these datasets often lack diversity in both content and distortion, and models trained on them generalize poorly to real-world conditions~\cite{ghadiyaram2017capture, nuutinen2016cvd2014, ying2021patch, sinno2018large}.

This limitation motivates a shift toward datasets with authentic distortions in user-generated content~\cite{hosu2017konstanz, sinno2018large, wang2024youtube, wang2019youtube}. 
Large-scale datasets like YouTube-UGC~\cite{wang2019youtube} and LSVQ~\cite{ying2021patch} aim to dramatically increase diversity in content, scene complexity, capture conditions, and device characteristics, thereby encouraging VQA models to learn more perceptually relevant and generalizable representations.   
Recently, the proliferation of AI-generated videos has created a new data regime, where distortions are often semantic or logical (\eg, with unrealistic objects or unnatural motion) rather than purely signal-level artifacts. 
Datasets like AIGVQA-DB~\cite{wang2025aigv} have been introduced to target these emerging distortions. However, even with ongoing dataset expansion, a persistent issue remains: a large fraction of available data may still be easy for existing VQA models, which risks inefficient use of labeling resources and limits the impact of additional annotations~\cite{sun2024analysis,wen2024modular}.

\section{Proposed Method: MDS-VQA}
\label{sec:method}
In this section, we present MDS-VQA, a model-informed dataset selection mechanism that prioritizes informative unlabeled videos for annotation to improve VQA efficiency and generalization (see Fig.~\ref{fig:overview}).

\subsection{Overview and Problem Formulation}

A productive development cycle for VQA should couple model improvement with data curation: models should be aware of where they fail, and new labels are then collected specifically to address those weaknesses. Motivated by this, we propose MDS-VQA, a model-informed data selection mechanism that prioritizes unlabeled videos most informative to the base VQA model. 

Let $\mathcal{U}$ denote a large unlabeled video pool (\eg, from a web crawl). Given a labeling budget, our goal is to select a subset $\mathcal{D}\subset \mathcal{U}$ for subjective annotation. MDS-VQA prioritizes videos that are both 1) difficult for the base VQA model and 2) diverse in content. Formally, we cast sample selection as a subset optimization problem~\cite{cao2024image}:
\begin{equation}
    \mathcal{D}=\operatorname*{arg max}_{\mathcal{S}\subset\mathcal{U}}\mathrm{Diff}(\mathcal{S};f)+\lambda\mathrm{Div}(\mathcal{S}),
\label{eq:framework}
\end{equation}
where $\mathcal{S}$ is a candidate video subset from the unlabeled pool $\mathcal{U}$ and $f(\cdot)$ is a top-performing VQA model. $\mathrm{Diff}(\cdot)$ measures how challenging the videos are for $f(\cdot)$, 
$\mathrm{Div}(\cdot)$ encourages content coverage, and $\lambda$ trades off the two terms.

After annotating the selected subset $\mathcal{D}$, we incorporate the newly labeled data to retrain or fine-tune the base VQA model, thus closing the loop between model diagnosis and data acquisition.

\subsection{Ranking-Based Difficulty Modeling}
\label{sec:learning2rank}

To instantiate $\mathrm{Diff}(\cdot)$, we augment the base VQA model $f(\cdot)$ with a failure predictor $g(\cdot)$ that estimates how likely a video is to expose the model's errors. In our main setting, $f(\cdot)$ is instantiated by VisualQuality-R1~\cite{wu2025visualquality} and $g(\cdot)$ is implemented by parameter-efficient low-rank adaptation (LoRA)~\cite{hu2022lora}, while keeping $f(\cdot)$ fixed.

\begin{table}[!t]
\centering
\renewcommand{\arraystretch}{1}
\caption{Structured text prompt used for training $g(\cdot)$.}
\label{tab:text prompt}
    \resizebox{1.0\linewidth}{!}{
    \begin{tabular}{p{\linewidth}}
    \toprule
    You are doing a video quality assessment task. \\
    Here is the question: Assess how difficult it is to evaluate this video's quality. The difficulty rating should be a float between $1$ and $5$, rounded to two decimal places, with $1$ representing very easy and $5$ representing very difficult. \\
    Please only output the final answer with only one score in the \texttt{<answer>} \texttt{</answer>} tags. \\
    \bottomrule
    \end{tabular}
}
\end{table}

\noindent\textbf{LoRA-Based Failure Predictor.} 
Concretely, we instantiate $g(\cdot)$ by attaching LoRA modules to the linear layers of the base model, while keeping the original weights fixed. For a linear projection with frozen weight $W_0 \in \mathbb{R}^{d_\mathrm{out}\times d_\mathrm{in}}$, LoRA parameterizes the mapping as
\begin{equation}
    W_\mathrm{LoRA} = W_0 +\Delta W = W_0 + \frac{\alpha}{r}BA,
\end{equation}
where $A\in \mathbb{R}^{r\times d_\mathrm{in}}$ and $B\in\mathbb{R}^{d_\mathrm{out}\times r}$ are trainable low-rank matrices, $r$ is the rank, and $\alpha$ is a scaling factor. During training, we use a fixed structured prompt that asks the model to output a difficulty rating in $[1,5]$ (higher means harder), formatted in \texttt{<answer>} tags (see Table~\ref{tab:text prompt}).

\noindent\textbf{Ranking-Based Training Objective.} 
An effective training objective is essential for reliably learning the failure predictor. Regressing absolute prediction error is sensitive to MOS scale differences across datasets, while binary ``easy \vs hard'' classification lacks granularity among difficult samples. We therefore formulate failure prediction as a learning-to-rank problem that only requires relative comparisons~\cite{cao2024image, yoo2019learning}. Given a training pair of videos $(x, y)$, the failure predictor outputs two scalars $g(x)$ and $g(y)$.
Under a Thurstone model~\cite{thurstone1927law}, we interpret these scores as the means of unit-variance Gaussian random variables, and compute the probability that video $x$ is more difficult than $y$ for the task of VQA as
\begin{equation}
    \hat{p}(x,y) = \Phi\left(\frac{g(x)-g(y)}{\sqrt{2}}\right),
\end{equation}
where $\Phi(\cdot)$ is the standard Gaussian cumulative distribution function. We construct supervisory signals from $f(\cdot)$'s predicted errors:
\begin{equation}
    p(x,y)=\left\{\begin{aligned} & 1&&\mathrm{if}\ |f(x)-\mu(x)|\geq|f(y)-\mu(y)|,\\ & 0&&\mathrm{otherwise},\end{aligned}\right.
\end{equation}
where $\mu(x)$ and $\mu(y)$ denote the MOSs of $x$ and $y$, respectively. We then optimize $g(\cdot)$ using the fidelity loss~\cite{tsai2007frank}:
\begin{equation}
\begin{split}
    \ell(x,y,p)=~&1-\sqrt{p(x,y)\hat{p}(x,y) } \\
    & - \sqrt{(1-p(x,y))(1-\hat{p}(x,y))},
\end{split}
\end{equation}
which encourages $g(\cdot)$ to assign larger scores to videos that induce greater predicted errors in $f(\cdot)$.

\subsection{Model-Informed Selection with Diversity}
\label{sec:sample selction}

After training $g(\cdot)$, we combine its difficulty estimates with a content diversity measure to select a maximally informative subset $\mathcal{D}\subset\mathcal{U}$, which is then annotated and subsequently used for active fine-tuning.

\noindent\textbf{Set Difficulty.} For a candidate subset $\mathcal{S}$, we quantify its difficulty by the mean predicted failure score:
\begin{equation}
    \mathrm{Diff}(\mathcal{S})=\frac{1}{\vert\mathcal{S}\vert}\sum_{x\in\mathcal{S}}g(x),
\end{equation}
where we omit the explicit dependence on $f(\cdot)$ in Eq.~\eqref{eq:framework} for notation simplicity.
A larger value indicates that the base VQA model struggles more on videos in $\mathcal{S}$, making them potentially more informative for improving generalization.

\noindent\textbf{Set Diversity.} 
To encourage broad coverage, we represent each video $x$ as a set of frame-level semantic features $\mathcal{F}_x$, extracted using the CLIP vision encoder~\cite{radford2021learning}. We then measure dissimilarity between two videos via the Chamfer distance between their frame-level features, capturing semantic variations beyond a single pooled descriptor:
\begin{equation}
\begin{aligned}
d_{\mathrm{CD}}(\mathcal{F}_x,\mathcal{F}_y)
&=\frac{1}{|\mathcal{F}_x|}\sum_{u\in \mathcal{F}_x}\min_{v\in \mathcal{F}_y}\|u-v\|_2^2 \\
&\quad +\frac{1}{|\mathcal{F}_y|}\sum_{v\in\mathcal{F}_y}\min_{u\in \mathcal{F}_x}\|v-u\|_2^2 .
\end{aligned}
\end{equation}
The diversity of a subset $\mathcal{S}$ is then defined
as the mean pairwise Chamfer distance:
\begin{equation}
    \mathrm{Div}(\mathcal{S})=\frac{1}{\binom{|\mathcal{S}|}{2}}\sum_{(x,y)\in\mathcal{S}} d_\mathrm{CD}(\mathcal{F}_{x},\mathcal{F}_{y}).
    \label{eq:Div}
\end{equation} 
Maximizing $\mathrm{Div}(\mathcal{S})$ encourages the selected videos to span a wide spectrum of scene types.

\noindent\textbf{Greedy Approximation.} 
Problem~\eqref{eq:framework} is combinatorial and NP-hard~\cite{natarajan1995sparse, davis1997adaptive}, so we adopt a greedy selection strategy. Starting from $\mathcal{D}_0 = \emptyset$, we iteratively add the video that best balances 1) its predicted difficulty and 2) its dissimilarity to already selected samples. 
At iteration $k+1$, we choose 
\begin{equation}
    x^{\star} = \underset{x \in \mathcal{U} \setminus \mathcal{D}_{k}}{\operatorname*{arg max}} \biggl( g(x) + \frac{\lambda}{\vert\mathcal{D}_{k}\vert}\sum_{y\in\mathcal{D}_{k}} d_\mathrm{CD} (\mathcal{F}_x, \mathcal{F}_{y} ) \biggr),
    \label{eq:sampling}
\end{equation}
and update $\mathcal{D}_{k+1} = \mathcal{D}_k\cup\{x^\star\}$ until the labeling budget is reached. Here, $\lambda$ is the same trade-off parameter.

By explicitly favoring samples that are both hard and non-redundant, MDS-VQA creates a targeted subset that more comprehensively reflects the base model's blind spots, leading to more efficient active fine-tuning.

\begin{table*}[t]
\centering
\caption{Failure identification performance. We compare MDS-VQA with eight competing methods on simulated unlabeled pools $\mathcal{U}$ from CGVDS, LIVE-Livestream, YouTube-SFV SDR, YouTube-SFV HDR2SDR, and AIGVQA-DB. Each entry reports SRCC/PLCC between the base model predictions and MOSs on the samples selected by each method;
lower values indicate stronger failure identification. Best results are highlighted in bold.
}
\resizebox{\textwidth}{!}{
\begin{tabular}{l|cccccc}
\toprule
\multicolumn{1}{l|}{\makebox[3.6cm][l]{Method}} &
\makebox[2.32cm][c]{CGVDS~\cite{zadtootaghaj2020quality}} &
  \makebox[2.2cm][c]{\begin{tabular}[c]{@{}c@{}}LIVE-\\ Livestream~\cite{shang2021assessment}\end{tabular}} &
  \makebox[2.3cm][c]{\begin{tabular}[c]{@{}c@{}}YouTube-SFV\\ SDR~\cite{wang2024youtube}\end{tabular}} &
  \makebox[2.3cm][c]{\begin{tabular}[c]{@{}c@{}}YouTube-SFV\\ HDR2SDR~\cite{wang2024youtube}\end{tabular}} &
  \makebox[3.cm][c]{AIGVQA-DB~\cite{wang2025aigv}} &
  \multicolumn{1}{|c}{\makebox[2.2cm][c]{Mean}} \\ 
\cmidrule(r){1-7}
\multicolumn{1}{l|}{Base model~\cite{wu2025visualquality}}       & 0.544 / 0.635 & 0.473 / 0.493 & 0.665 / 0.710 & 0.538 / 0.591 & 0.733 / 0.740 & \multicolumn{1}{|c}{0.591 / 0.634} \\ 
\cmidrule(r){1-7}
\multicolumn{1}{l|}{Random sampling}                           & 0.673 / 0.782 & 0.521 / 0.555 & 0.642 / 0.787 & 0.438 / 0.407 & 0.652 / 0.729 & \multicolumn{1}{|c}{0.585 / 0.652} \\
\multicolumn{1}{l|}{Core-set selection~\cite{sener2017active}} & 0.415 / 0.599 & 0.289 / 0.378 & 0.599 / 0.741 & 0.516 / 0.555 & 0.676 / 0.742 & \multicolumn{1}{|c}{0.499 / 0.603} \\
\multicolumn{1}{l|}{Sampling by RD~\cite{wu2018pool}}          & 0.411 / 0.412 & 0.253 / 0.352 & 0.495 / 0.535 & 0.372 / 0.592 & 0.640 / 0.660 & \multicolumn{1}{|c}{0.434 / 0.510} \\
\multicolumn{1}{l|}{MC dropout~\cite{pop2018deep}}             & 0.379 / 0.440 & 0.333 / 0.350 & 0.411 / 0.452 & 0.359 / 0.490 & 0.754 / 0.757 & \multicolumn{1}{|c}{0.447 / 0.498} \\
\multicolumn{1}{l|}{Greedy sampling~\cite{bhaskara2019greedy}} & 0.331 / 0.564 & 0.461 / 0.586 & 0.472 / 0.532 & 0.334 / 0.451 & 0.659 / 0.663 & \multicolumn{1}{|c}{0.451 / 0.559} \\
\multicolumn{1}{l|}{ALCS~\cite{yan2022clustering}}             & 0.388 / 0.817 & 0.310 / 0.545 & 0.485 / 0.696 & 0.342 / 0.614 & 0.597 / 0.600 & \multicolumn{1}{|c}{0.424 / 0.654} \\
\multicolumn{1}{l|}{FreeSel~\cite{xie2023towards}}             & 0.252 / 0.450 & 0.232 / 0.418 & 0.546 / 0.690 & 0.262 / 0.422 & 0.565 / 0.643 & \multicolumn{1}{|c}{0.371 / 0.525} \\
\multicolumn{1}{l|}{NoiseStability~\cite{li2024deep}}         & 0.362 / 0.375 & 0.267 / 0.493 & 0.326 / 0.661 & 0.274 / 0.485 & 0.597 / 0.661 & \multicolumn{1}{|c}{0.365 / 0.535} \\
\cmidrule(r){1-7}
\multicolumn{1}{l|}{{MDS-VQA} (Ours)}                          & \textbf{0.162} / \textbf{0.316} & \textbf{0.133} / \textbf{0.288} & \textbf{0.264} / \textbf{0.361} & \textbf{0.161} / \textbf{0.354} & \textbf{0.487} / \textbf{0.487} & \multicolumn{1}{|c}{\textbf{0.241} / \textbf{0.361}} \\
\bottomrule
\end{tabular}
}
\label{tab:failure identification}
\end{table*}

\subsection{Subset Labeling and Active Fine-Tuning}
\label{subsec:lft}

\noindent\textbf{Subset Labeling.} 
After obtaining the budgeted subset $\mathcal{D}\subset \mathcal{U}$ using Eq.~\eqref{eq:sampling}, we conduct subjective experiments to acquire reliable human quality judgments for the selected videos using a standard protocol (\eg, absolute category rating or paired comparison). Importantly, we represent the newly annotated data in a \textit{scale-free} pairwise form by constructing labeled pairs from the annotated videos when needed. This pairwise formulation is invariant to the absolute scoring scale of any individual study, allowing the newly labeled subset to be merged seamlessly with existing pairwise VQA datasets without cross-dataset perceptual scale alignment~\cite{9369977, wang2021active, wang2021troubleshooting}.

\noindent\textbf{Active Fine-Tuning with LoRA.} 
We then update the base VQA model $f(\cdot)$ using both existing VQA data and the newly labeled subset. Our fine-tuning procedure follows the training recipe of VisualQuality-R1 (\ie, prompting, on-the-fly pair construction within batches, and reinforcement-learning-to-rank optimization), 
but with one key modification: we replace full fine-tuning with parameter-efficient LoRA, as stated in Sec.~\ref{sec:learning2rank}. This design mitigates overfitting or catastrophic forgetting while still facilitating efficient adaptation to model-challenging samples.

\noindent\textbf{Iteration (Optional).} 
The updated VQA model can be used to re-estimate difficulty and repeat Secs.~\ref{sec:learning2rank}-\ref{subsec:lft} for additional rounds of selection when labeling budget permits, progressively shifting annotation effort toward the evolving failure modes of the VQA model.

\section{Experiments}
\label{sec:experiment}

In this section, we assess MDS-VQA from three complementary angles: 1) failure identification (\ie, how effectively MDS-VQA surfaces samples on which the base VQA model disagrees with human judgments), 2) active fine-tuning gains (\ie, how much performance improves after fine-tuning on the selected subset), and 3) generalization under gMAD competition (\ie, worst-case model differentiation beyond average-case correlation metrics).

\subsection{Experimental Setups}

\noindent\textbf{Datasets.} 
We use five VQA datasets that cover diverse video application scenarios: YouTube-UGC~\cite{wang2019youtube}, CGVDS~\cite{zadtootaghaj2020quality}, LIVE-Livestream~\cite{shang2021assessment}, YouTube-SFV+HDR~\cite{wang2024youtube}, and AIGVQA-DB~\cite{wang2025aigv}. 
YouTube-UGC comprises approximately $1,500$ user-generated video clips with authentic distortions (\eg, blockiness, blur, and jerkiness) and is used as the source dataset for training the base model.
The other four datasets (in total over $42,000$ videos) serve as target-domain ``unlabeled'' pools. CGVDS~\cite{zadtootaghaj2020quality} targets cloud-streamed gaming content encoded by hardware-accelerated H.264/MPEG-AVC at 
$60$ frames per second under various bitrates and resolutions. LIVE-Livestream~\cite{shang2021assessment} focuses on 4K live streaming of professionally generated sports content, featuring characteristic impairments arising from capture artifacts and adverse network conditions. YouTube-SFV+HDR~\cite{wang2024youtube} contains short-form standard- and high-dynamic-range 1080p videos across popular categories (\eg, dancing and cooking) with typical streaming distortions. AIGVQA-DB~\cite{wang2025aigv} benchmarks AI-generated videos produced by text-to-video models, capturing unique artifacts such as unrealistic objects and unnatural motion.

For worst-case evaluation, we additionally use LSVQ-1080p~\cite{ying2021patch} (with $3,573$ high-resolution videos) to conduct gMAD testing, as it contains high-quality, subtle perceptual differences that are challenging for generalization.

\begin{table*}[!t]
\centering
\caption{Active fine-tuning performance. Models are fine-tuned on the YouTube-UGC training set and subsets selected by different methods under a $5\%$ labeling budget per target domain. 
Each entry reports SRCC/PLCC on the corresponding test set.
}
\resizebox{\textwidth}{!}{
\begin{tabular}{l|c|cccccc}
\toprule
\multicolumn{1}{l|}{Method} & \multicolumn{1}{c|}{\begin{tabular}[c]{@{}c@{}}YouTube-\\ UGC~\cite{wang2019youtube}\end{tabular}} &
  CGVDS~\cite{zadtootaghaj2020quality} &
  \begin{tabular}[c]{@{}c@{}}LIVE-\\ Livestream~\cite{shang2021assessment}\end{tabular} &
  \begin{tabular}[c]{@{}c@{}}YouTube-SFV\\ SDR~\cite{wang2024youtube}\end{tabular} &
  \begin{tabular}[c]{@{}c@{}}YouTube-SFV\\ HDR2SDR~\cite{wang2024youtube}\end{tabular} &
  AIGVQA-DB~\cite{wang2025aigv} &  
  \multicolumn{1}{|c}{Mean} \\ 
\cmidrule(r){1-8}
\multicolumn{1}{l|}{Base model~\cite{wu2025visualquality}}       & \multicolumn{1}{c|}{0.708 / 0.709} & 0.766 / 0.780 & 0.561 / 0.587 & 0.666 / 0.718 & 0.495 / 0.557 & 0.711 / 0.748 & \multicolumn{1}{|c}{0.651 / 0.683} \\ 
\cmidrule(r){1-8}
\multicolumn{1}{l|}{Random sampling}                           & \multicolumn{1}{c|}{0.760 / 0.756} & 0.807 / 0.804 & 0.569 / 0.628 & 0.703 / 0.761 & \textbf{0.518} / 0.588 & 0.756 / 0.751 & \multicolumn{1}{|c}{0.686 / 0.715} \\
\multicolumn{1}{l|}{Core-set selection~\cite{sener2017active}} & \multicolumn{1}{c|}{0.776 / 0.765} & 0.832 / 0.842 & 0.601 / 0.610 & 0.700 / 0.760 & 0.517 / 0.593 & 0.788 / 0.787 & \multicolumn{1}{|c}{0.702 / 0.726} \\
\multicolumn{1}{l|}{Sampling by RD~\cite{wu2018pool}}          & \multicolumn{1}{c|}{0.795 / 0.784} & 0.780 / 0.825 & 0.597 / 0.602 & 0.729 / 0.778 & 0.517 / 0.587 & 0.772 / 0.767 & \multicolumn{1}{|c}{0.698 / 0.724} \\
\multicolumn{1}{l|}{MC dropout~\cite{pop2018deep}}             & \multicolumn{1}{c|}{0.760 / 0.753} & 0.819 / 0.831 & 0.595 / 0.609 & 0.710 / 0.764 & 0.508 / 0.585 & 0.758 / 0.764 & \multicolumn{1}{|c}{0.692 / 0.718} \\
\multicolumn{1}{l|}{Greedy sampling~\cite{bhaskara2019greedy}} & \multicolumn{1}{c|}{0.781 / 0.773} & 0.831 / 0.834 & 0.585 / 0.605 & 0.711 / 0.774 & 0.500 / 0.580 & 0.775 / 0.777 & \multicolumn{1}{|c}{0.697 / 0.724} \\
\multicolumn{1}{l|}{ALCS~\cite{yan2022clustering}}             & \multicolumn{1}{c|}{0.785 / 0.772} & 0.807 / 0.823 & 0.626 / 0.636 & 0.727 / 0.781 & 0.513 / 0.579 & 0.789 / \textbf{0.792}& \multicolumn{1}{|c}{0.708 / 0.730} \\
\multicolumn{1}{l|}{FreeSel~\cite{xie2023towards}}             & \multicolumn{1}{c|}{0.814 / 0.798} & 0.832 / 0.849 & 0.627 / 0.646 & 0.719 / 0.787 & 0.498 / 0.590 & 0.789 / 0.785 & \multicolumn{1}{|c}{0.713 / 0.742} \\
\multicolumn{1}{l|}{NoiseStability~\cite{li2024deep}}         & \multicolumn{1}{c|}{0.806 / 0.798} & 0.826 / 0.847 & 0.601 / 0.616 & 0.727 / 0.780 & 0.500 / 0.594 & \textbf{0.790} / 0.786 & \multicolumn{1}{|c}{0.708 / 0.737} \\
\cmidrule(r){1-8}
\multicolumn{1}{l|}{MDS-VQA (Ours)}                          & \multicolumn{1}{c|}{\textbf{0.819} / \textbf{0.807}} & \textbf{0.874} / \textbf{0.875} & \textbf{0.632} / \textbf{0.654} & \textbf{0.731} / \textbf{0.794} & 0.507 / \textbf{0.595} & 0.769 / 0.769 & \multicolumn{1}{|c}{\textbf{0.722} / \textbf{0.749}} \\
\bottomrule
\end{tabular}
}
\label{tab:active finetune}
\end{table*}

\noindent\textbf{Implementation Details.} 
\label{sec:implementation details}
We compare the proposed MDS-VQA against a broad set of data selection strategies, including random sampling, core-set selection~\cite{sener2017active}, sampling by representativeness diversity (RD)~\cite{wu2018pool}, Monte Carlo (MC) dropout~\cite{pop2018deep}, greedy sampling~\cite{bhaskara2019greedy}, ALCS~\cite{yan2022clustering}, FreeSel~\cite{xie2023towards}, and NoiseStability~\cite{li2024deep}. 

We first train VisualQuality-R1~\cite{wu2025visualquality} on YouTube-UGC to obtain the base quality model. VisualQuality-R1 is built on Qwen2.5-VL~\cite{bai2025qwen2}, a video-pretrained backbone that supports temporal modeling through dynamic frame sampling and time-aligned embeddings, making it well suited for video-based quality reasoning and scoring. We then train the failure predictor $g(\cdot)$ by attaching LoRA modules to the base VQA model, using rank $r=64$, scaling $\alpha = 128$, and dropout $p=0.05$. Training is carried out by AdamW~\cite{loshchilov2017decoupled} with learning rate $1\times10^{-5}$ followed by a linear decay schedule and batch size $8$ for $10$ epochs~\cite{shen2025vlm}.

After learning $g(\cdot)$, we select informative samples from target-domain unlabeled pools using the greedy rule in Eq.~\eqref{eq:sampling} with $\lambda=0.25$, under a fixed labeling budget of $5\%$ per target dataset. Finally, we actively fine-tune the base model on the union of the full YouTube-UGC training set and the selected subsets from target domains. We use the same hyperparameter configuration as in training the failure predictor (see Supplementary for additional details).

\subsection{Main Results}

\noindent\textbf{Failure Identification Results.} We first evaluate whether a selection method can reliably surface failure cases of the base VQA model. Concretely, after each method selects a $5\%$ subset from a target-domain pool, we compute the SRCC and Pearson linear correlation coefficient (PLCC) between the base model predictions and MOSs on the selected subset. A lower correlation number indicates a more failure-focused subset as it reflects stronger disagreement with human judgments.

As shown in Table~\ref{tab:failure identification}, across all evaluated target sets, MDS-VQA achieves the lowest (best) SRCC/PLCC results, indicating the strongest capability to identify difficult samples. For example, on CGVDS, MDS-VQA reduces SRCC to $0.162$ compared with $0.673$ for random sampling. The same pattern holds for live streaming, short-form streaming, and text-to-video generation, where MDS-VQA attains substantially lower SRCC values than alternatives. 
Notably, this advantage emerges even though neither the base VQA model nor the failure predictor sees target-domain data and labels. While the quality mapping learned by the base model may shift under domain changes, the resulting uncertainty and inconsistency patterns are more domain-agnostic and thus tend to transfer. This explains why MDS-VQA can better surface failure cases from target pools than either uncertainty-only~\cite{pop2018deep} or diversity-only~\cite{sener2017active, wu2018pool, bhaskara2019greedy, yan2022clustering, xie2023towards} selection.

\begin{figure*}[!t]
	\centering
	\includegraphics[width=\textwidth]{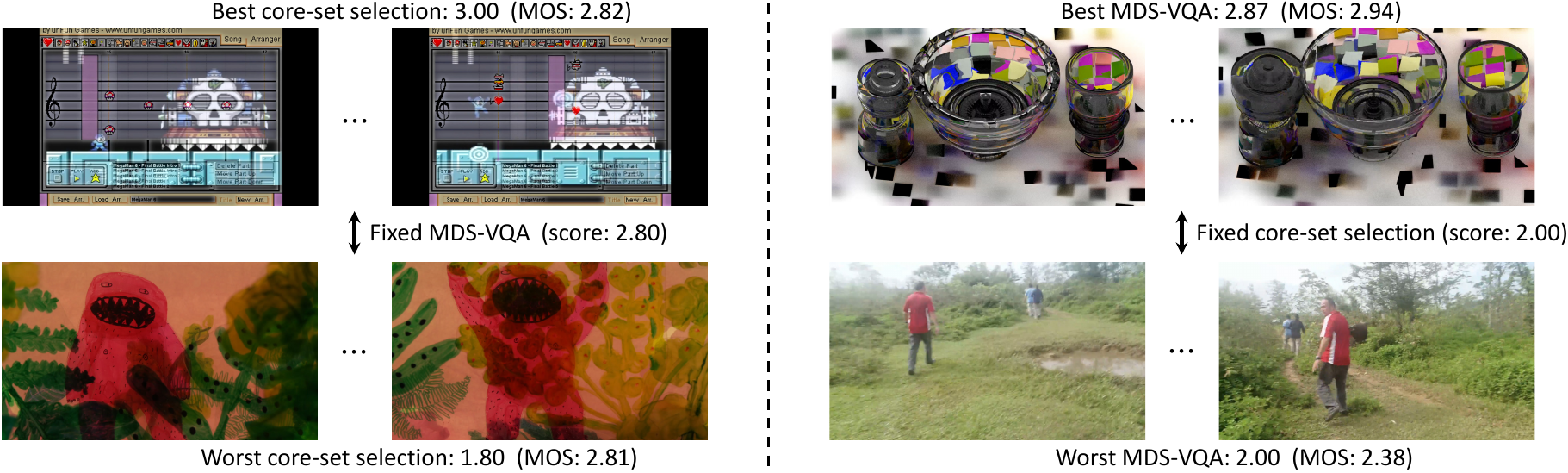}
	\caption{Representative gMAD pairs between VQA models induced by MDS-VQA and core-set selection~\cite{sener2017active}. Left: gMAD pairs found by fixing MDS-VQA-induced model predictions and searching for videos that maximally differentiate the core-set-induced model. Right: the reverse setting. Predicted scores and MOSs are shown for each pair, both on a $[1,5]$ scale where higher values indicate better predicted and perceived quality, respectively.
    }
	\label{fig:gMAD}
\end{figure*}

\noindent\textbf{Active Fine-Tuning Results.}
Next, we test whether the selected subsets are actually useful for improving the base VQA model. We fine-tune the model on the YouTube-UGC training set~\cite{wang2019youtube} and the selected $5\%$ subsets from target domains, and evaluate performance across test splits of all datasets. From Table~\ref{tab:active finetune}, we find that MDS-VQA yields the highest mean correlation, suggesting that the selected samples provide broad and transferable supervision rather than narrow domain-specific overfitting. We attribute these gains in part to explicitly enforcing \textit{content diversity} during selection, which avoids repeatedly labeling near-duplicate hard cases and encourages coverage of complementary failure modes across varied content and distortion conditions (see also the results in Table~\ref{tab:lambda}).

\begin{table}[!t]
\centering
\caption{Quality prediction performance comparison under SRCC and gMAD competition~\cite{ma2018group}. We report each method’s overall SRCC rank (based on the mean results in Table~\ref{tab:active finetune}) and gMAD rank (evaluated on LSVQ-1080p~\cite{ying2021patch}). $\Delta$~Rank denotes the difference between the two rankings. 
}
\resizebox{\linewidth}{!}{
\begin{tabular}{lccc}
\toprule
Method & 
\makebox[2.cm][c]{SRCC Rank} & 
\makebox[2.cm][c]{gMAD Rank} & 
\makebox[1.4cm][c]{$\Delta$ Rank} \\
\cmidrule(r){1-4}
MDS-VQA (Ours)                            & 1 & 1 & 0  \\ 
\cmidrule(r){1-4}
FreeSel~\cite{xie2023towards}             & 2 & 2 & 0   \\
NoiseStability~\cite{li2024deep}         & 3 & 6 & -3  \\
ALCS~\cite{yan2022clustering}             & 4 & 5 & -1  \\
Core-set selection~\cite{sener2017active} & 5 & 8 & -3  \\
Sampling by RD~\cite{wu2018pool}          & 6 & 3 & 3   \\
Greedy sampling~\cite{bhaskara2019greedy} & 7 & 4 & 3   \\
MC dropout~\cite{pop2018deep}             & 8 & 9 & -1  \\
Random sampling                           & 9 & 10 & -1 \\
Base model~\cite{wu2025visualquality}       & 10 & 7 & 3  \\
\bottomrule
\end{tabular}
}
\label{tab:gMAD}
\end{table}

\noindent\textbf{gMAD Competition Results.}
Average-case correlation can hide rare but consequential failures. We therefore conduct a gMAD competition~\cite{ma2018group} on LSVQ-1080p~\cite{ying2021patch}, which explicitly searches for samples that maximally distinguish two models and thus probes worst-case generalization. From Table~\ref{tab:gMAD}, we observe that MDS-VQA attains the top gMAD rank (and also the top SRCC rank), indicating that its gains are consistent under both average-case and worst-case evaluation criteria. In contrast, several competing methods exhibit noticeable SRCC-gMAD rank discrepancies. The qualitative gMAD analysis in Fig.~\ref{fig:gMAD} further illustrates this generalization. When MDS-VQA is the attacker, it reveals that a core-set-induced model can severely under-score videos with animation and abstract patterns despite high MOSs, whereas the MDS-VQA-induced model remains more consistent with human perception when attacked by core-set selection~\cite{sener2017active}.

\subsection{Ablation Studies}

We perform ablations to isolate the contributions of the key design components. Unless otherwise stated, results are reported on YouTube-SFV SDR~\cite{wang2024youtube}, with base models trained on YouTube-UGC~\cite{wang2019youtube} under the default settings.

\begin{table}[!t]
\centering
\caption{Ablation on training losses for the failure predictor $g(\cdot)$. The default setting is highlighted in bold.}
\label{tab:loss function}
\resizebox{\linewidth}{!}{
\sbox{\mybox}{PLCC\qquad}
\begin{tabular}{lcccc}
\toprule
\multicolumn{1}{l|}{}                 & \multicolumn{2}{c|}{Failure identification} & \multicolumn{2}{c}{Active fine-tuning} \\
\multicolumn{1}{l|}{\multirow{-2}{*}{\makebox[3.1cm][l]{Loss}}} & \makebox[.8\wd\mybox][c]{SRCC} & \multicolumn{1}{c|}{\makebox[.1\wd\mybox][c]{PLCC}} & \makebox[.7\wd\mybox][c]{SRCC} & \makebox[.1\wd\mybox][c]{PLCC} \\
\cmidrule(r){1-5}
\multicolumn{1}{l|}{Base model~\cite{wu2025visualquality}}         & 0.665 & \multicolumn{1}{c|}{0.710} & 0.666 & 0.718 \\
\cmidrule(r){1-5}
\multicolumn{1}{l|}{Removing $g(\cdot)$}  & 0.509 & \multicolumn{1}{c|}{0.605} & 0.705 & 0.770 \\
\multicolumn{1}{l|}{Classification loss} & 0.415 & \multicolumn{1}{c|}{0.482} & 0.715 & 0.771 \\
\multicolumn{1}{l|}{Regression loss} & 0.396 & \multicolumn{1}{c|}{0.472} & 0.714 & 0.780 \\
\multicolumn{1}{l|}{\textbf{Ranking loss}} & \textbf{0.264} & \multicolumn{1}{c|}{\textbf{0.361}} & \textbf{0.726} & \textbf{0.789} \\
\bottomrule
\end{tabular}
}
\end{table}

\noindent\textbf{Effect of Failure Prediction Losses.} We compare ranking-based training of the failure predictor $g(\cdot)$ against classification (with the cross-entropy loss) and regression (with the mean squared error), and also include a diversity-only variant that removes $g(\cdot)$ from selection. From Table~\ref{tab:loss function}, we see that the ranking loss consistently yields the strongest failure identification and the largest gains after active fine-tuning. This behavior is expected because our goal is inherently \textit{ordinal}: we only need $g(\cdot)$ to induce a reliable relative ordering of samples by model difficulty, rather than to predict a calibrated absolute failure magnitude. In contrast, regression losses implicitly assume a stable numeric scale and can be dominated by a few high-error outliers or label noise. Classification further compresses supervision into coarse bins, discarding within-bin difficulty structure that is crucial for selecting the top fraction under a fixed budget. Additionally, ranking objectives are less sensitive to dataset-specific score ranges and heteroscedasticity, making them better aligned with cross-domain selection where the quality scale and error distribution may shift. Finally, the diversity-only variant underperforms because semantic coverage alone does not target the base model's blind spots, underscoring the importance of explicit difficulty modeling.  

\begin{table}[!t]
\centering
\caption{Ablation on the difficulty-diversity trade-off weight $\lambda$.}
\label{tab:lambda}
\resizebox{\linewidth}{!}{
\sbox{\mybox}{PLCC\qquad}
\begin{tabular}{lcccc}
\toprule
\multicolumn{1}{l|}{}                 & \multicolumn{2}{c|}{Failure identification} & \multicolumn{2}{c}{Active fine-tuning} \\
\multicolumn{1}{l|}{\multirow{-2}{*}{\makebox[3.1cm][l]{Weight}}} & \makebox[.8\wd\mybox][c]{SRCC} & \multicolumn{1}{c|}{\makebox[.1\wd\mybox][c]{PLCC}} & \makebox[.7\wd\mybox][c]{SRCC} & \makebox[.1\wd\mybox][c]{PLCC} \\
\cmidrule(r){1-5}
\multicolumn{1}{l|}{Base model~\cite{wu2025visualquality}}        & 0.665 & \multicolumn{1}{c|}{0.710} & 0.666 & 0.718 \\
\cmidrule(r){1-5}
\multicolumn{1}{l|}{$\lambda=0$}     & 0.317 & \multicolumn{1}{c|}{0.385} & 0.715 & 0.781 \\
\multicolumn{1}{l|}{$\lambda=0.125$} & 0.295 & \multicolumn{1}{c|}{0.367} & 0.720 & 0.784 \\
\multicolumn{1}{l|}{\rlap{$\boldsymbol{\lambda}$}\phantom{$\lambda$}$\mathbf{\;=0.250}$} & \textbf{0.264} & \multicolumn{1}{c|}{0.361} & \textbf{0.726} & \textbf{0.789} \\
\multicolumn{1}{l|}{$\lambda=0.500$} & 0.281 & \multicolumn{1}{c|}{\textbf{0.356}} & 0.725 & 0.784 \\
\bottomrule
\end{tabular}
}
\end{table}

\noindent\textbf{Effect of $\boldsymbol{\lambda}$ (Difficulty-Diversity Trade-Off).} 
We vary $\lambda$ in Eq.~\eqref{eq:sampling} to quantify the role of diversity. As shown in Table~\ref{tab:lambda}, introducing diversity steadily improves both failure identification and fine-tuning performance compared to $\lambda = 0$ (\ie, difficulty-only), with the best overall balance at $\lambda = 0.25$. Increasing $\lambda$ further to $0.5$ slightly degrades performance, indicating that over-emphasizing coverage can dilute the ``hardness'' signal and reduce informativeness.

\noindent\textbf{Effect of Diversity Features.} 
We evaluate CLIP features~\cite{radford2021learning} against those derived from SigLIP 2~\cite{tschannen2025siglip} as well as internal representations extracted from the base VQA model, VisualQuality-R1~\cite{wu2025visualquality}, for computing the diversity term. As detailed in Table~\ref{tab:diversity features}, CLIP and SigLIP 2 yield comparable results, indicating that MDS-VQA is largely robust to the specific choice of diversity representations as long as they are generic visual embeddings trained at scale. In contrast, VisualQuality-R1 features perform the worst, likely because quality-oriented fine-tuning biases the representations toward perceptual degradations and reduces their ability to capture broad semantic variations for coverage. Given its widespread adoption and comparable performance to SigLIP 2, we use CLIP as a practical default.

\begin{table}[!t]
\centering
\caption{Ablation on diversity feature representations.}
\label{tab:diversity features}
\resizebox{\linewidth}{!}{
\sbox{\mybox}{PLCC\qquad}
\begin{tabular}{lcccc}
\toprule
\multicolumn{1}{l|}{}                 & \multicolumn{2}{c|}{Failure identification} & \multicolumn{2}{c}{Active fine-tuning} \\
\multicolumn{1}{l|}{\multirow{-2}{*}{Diversity feature}} & \makebox[.8\wd\mybox][c]{SRCC} & \multicolumn{1}{c|}{\makebox[.1\wd\mybox][c]{PLCC}} & \makebox[.7\wd\mybox][c]{SRCC} & \makebox[.1\wd\mybox][c]{PLCC} \\
\cmidrule(r){1-5}
\multicolumn{1}{l|}{Base model~\cite{wu2025visualquality}}         & 0.665 & \multicolumn{1}{c|}{0.710} & 0.666 & 0.718 \\
\cmidrule(r){1-5}
\multicolumn{1}{l|}{VisualQuality-R1~\cite{wu2025visualquality}}  & 0.397 & \multicolumn{1}{c|}{0.452} & 0.720 & 0.783 \\
\multicolumn{1}{l|}{\textbf{CLIP}~\cite{radford2021learning}} & \textbf{0.264} & \multicolumn{1}{c|}{0.361} & 0.726 & 0.789 \\
\multicolumn{1}{l|}{SigLIP 2~\cite{tschannen2025siglip}} & 0.279 & \multicolumn{1}{c|}{\textbf{0.350}} & \textbf{0.730} & \textbf{0.791} \\
\bottomrule
\end{tabular}
}
\end{table}

\begin{table}[!t]
\centering
\caption{Ablation on the LoRA rank $r$ for the failure predictor.}
\label{tab:lora rank}
\resizebox{\linewidth}{!}{
\sbox{\mybox}{PLCC\qquad}
\begin{tabular}{lcccc}
\toprule
\multicolumn{1}{l|}{}                 & \multicolumn{2}{c|}{Failure identification} & \multicolumn{2}{c}{Active fine-tuning} \\
\multicolumn{1}{l|}{\multirow{-2}{*}{\makebox[3.1cm][l]{LoRA rank}}} & \makebox[.8\wd\mybox][c]{SRCC} & \multicolumn{1}{c|}{\makebox[.1\wd\mybox][c]{PLCC}} & \makebox[.7\wd\mybox][c]{SRCC} & \makebox[.1\wd\mybox][c]{PLCC} \\
\cmidrule(r){1-5}
\multicolumn{1}{l|}{$r$ = 8} & 0.297 & \multicolumn{1}{c|}{0.425} & \textbf{0.727} & 0.788 \\
\multicolumn{1}{l|}{\textbf{$\mathbf{\textit{r}}$ = 64}} & \textbf{0.264} & \multicolumn{1}{c|}{\textbf{0.361}} & 0.726 & \textbf{0.789} \\
\multicolumn{1}{l|}{$r$ = 128} & 0.273 & \multicolumn{1}{c|}{0.418} & 0.724 & \textbf{0.789} \\
\bottomrule
\end{tabular}
}
\end{table}

\noindent\textbf{Effect of LoRA Rank.}
We next ablate the LoRA rank $r$ used to instantiate the failure predictor in Table~\ref{tab:lora rank}. Setting a small rank $r=8$ weakens failure identification, indicating insufficient capacity to learn reliable difficulty ordering. The default $r=64$ achieves the best failure identification. Increasing to $r=128$ degrades performance, suggesting mild overfitting or reduced cross-domain generalization with an over-parameterized adapter. Active fine-tuning, on the other hand, is largely stable across ranks, so we adopt $r=64$ as the best accuracy-efficiency trade-off.

\begin{table}[!t]
\centering
\caption{Ablation on base VQA models.}
\label{tab:architecture}
\resizebox{\columnwidth}{!}{
\sbox{\mybox}{PLCC\qquad}
\begin{tabular}{lcccc}
\toprule
\multicolumn{1}{l|}{}                   & \multicolumn{2}{c|}{Failure identification}      & \multicolumn{2}{c}{Active fine-tuning}      \\
\multicolumn{1}{l|}{\multirow{-2}{*}{Method}} & \makebox[.8\wd\mybox][c]{SRCC} & \multicolumn{1}{c|}{\makebox[.1\wd\mybox][c]{PLCC}} & \makebox[.7\wd\mybox][c]{SRCC} & \makebox[.1\wd\mybox][c]{PLCC} \\
\hline
\multicolumn{5}{l}{\cellcolor[HTML]{EFEFEF}\rule[-1.2ex]{0pt}{3.7ex}CNN-based UVQ~\cite{wang2021rich}}                                                                 \\
\multicolumn{1}{l|}{\rule[0ex]{0pt}{2.4ex}Base model~\cite{wang2021rich}} & 0.590 & \multicolumn{1}{c|}{0.644} & 0.577 & 0.648 \\ 
\cmidrule(r){1-5}
\multicolumn{1}{l|}{Random sampling}    & 0.577 & \multicolumn{1}{c|}{0.632} & 0.591 & 0.664 \\
\multicolumn{1}{l|}{Core-set selection~\cite{sener2017active}} & 0.417 & \multicolumn{1}{c|}{0.389} & 0.625 & 0.707 \\
\multicolumn{1}{l|}{Sampling by RD~\cite{wu2018pool}}     & 0.451 & \multicolumn{1}{c|}{0.510} & 0.623 & 0.716 \\
\multicolumn{1}{l|}{MC dropout~\cite{pop2018deep}}        & 0.278 & \multicolumn{1}{c|}{0.374} & 0.636 & 0.719 \\
\multicolumn{1}{l|}{Greedy sampling~\cite{bhaskara2019greedy}}    & 0.345 & \multicolumn{1}{c|}{0.393} & 0.631 & 0.719 \\
\multicolumn{1}{l|}{ALCS~\cite{yan2022clustering}}        & 0.308 & \multicolumn{1}{c|}{0.408} & 0.640 & 0.705 \\
\multicolumn{1}{l|}{FreeSel~\cite{xie2023towards}}        & 0.462 & \multicolumn{1}{c|}{0.525} & 0.645 & 0.723 \\
\multicolumn{1}{l|}{NoiseStability~\cite{li2024deep}}    & 0.224 & \multicolumn{1}{c|}{0.305} & 0.661 & 0.720 \\
\cmidrule(r){1-5}
\multicolumn{1}{l|}{\rule[-1.1ex]{0pt}{3.0ex}MDS-VQA (Ours)}      & \textbf{0.101} & \multicolumn{1}{c|}{\textbf{0.137}} & \textbf{0.677} & \textbf{0.729} \\ 
\hline
\multicolumn{5}{l}{\cellcolor[HTML]{EFEFEF}\rule[-1.2ex]{0pt}{3.7ex}Transformer-based ModularBVQA~\cite{wen2024modular}}                                          \\
\multicolumn{1}{l|}{\rule[0ex]{0pt}{2.4ex}Base model~\cite{wen2024modular}} & 0.636 & \multicolumn{1}{c|}{0.732} & 0.573 & 0.706 \\ 
\cmidrule(r){1-5}
\multicolumn{1}{l|}{Random sampling}    & 0.606 & \multicolumn{1}{c|}{0.696} & 0.659 & 0.743 \\
\multicolumn{1}{l|}{Core-set selection~\cite{sener2017active}} & 0.269 & \multicolumn{1}{c|}{0.718} & 0.686 & 0.769 \\
\multicolumn{1}{l|}{Sampling by RD~\cite{wu2018pool}} & 0.548 & \multicolumn{1}{c|}{0.641} & 0.677 & 0.744 \\
\multicolumn{1}{l|}{MC dropout~\cite{pop2018deep}}         & 0.414 & \multicolumn{1}{c|}{0.454} & 0.680 & 0.758 \\
\multicolumn{1}{l|}{Greedy sampling~\cite{bhaskara2019greedy}} & 0.466 & \multicolumn{1}{c|}{0.453} & 0.674 & 0.761 \\
\multicolumn{1}{l|}{ALCS~\cite{yan2022clustering}} & 0.476 & \multicolumn{1}{c|}{0.550} & 0.673 & 0.762 \\
\multicolumn{1}{l|}{FreeSel~\cite{xie2023towards}} & 0.412 & \multicolumn{1}{c|}{0.711} & 0.689 & 0.769 \\
\multicolumn{1}{l|}{NoiseStability~\cite{li2024deep}} & 0.342 & \multicolumn{1}{c|}{0.402} & 0.694 & 0.769 \\
\cmidrule(r){1-5}
\multicolumn{1}{l|}{MDS-VQA (Ours)} & \textbf{0.124} & \multicolumn{1}{c|}{\textbf{0.384}} & \textbf{0.713} & \textbf{0.771} \\ 
\bottomrule
\end{tabular}
}
\end{table}

\noindent\textbf{Generality across Base VQA Models.} 
Finally, we apply MDS-VQA to two alternative VQA models with substantially different architectures: the CNN-based UVQ~\cite{wang2021rich} and the Transformer-based ModularBVQA~\cite{wen2024modular} using full fine-tuning. As shown in Table~\ref{tab:architecture}, in both cases, MDS-VQA achieves the best failure identification and active fine-tuning performance. These consistent gains across a reasoning-induced, prompt-based predictor and fully discriminative quality models suggest that the proposed ``hard-and-diverse'' selection principle underlying MDS-VQA is largely architecture-agnostic and can serve as a plug-and-play add-on for different VQA backbones.

\section{Conclusion and Future Work}
\label{sec:conclusion}

In this paper, we have presented MDS-VQA, a model-informed data selection mechanism that reconnects model-centric VQA development and data-centric dataset construction. MDS-VQA has been proven effective at identifying informative failure cases and facilitating active fine-tuning with a constrained labeling budget across multiple datasets and backbone architectures.

Looking forward, several promising and practical directions can further strengthen MDS-VQA. First, the current ranking-based failure predictor may yield \textit{uncalibrated} difficulty scores, making the difficulty-diversity trade-off less convenient to tune. An adaptive, self-calibrating mechanism would substantially improve usability across deployment settings. Second, difficulty modeling can be enriched beyond a single scalar by incorporating fine-grained failure taxonomy (\eg, spatial artifacts, temporal consistency, and semantic plausibility) so the selection process can target which weaknesses to fix. Third, diversity can be refined with spatiotemporal and motion-aware representations (and potentially multi-modal cues such as audio and metadata) to avoid selecting clips that are semantically different yet perceptually redundant, thereby improving coverage of distinct failure modes at the same labeling budget. Finally, a practical next step is to operationalize MDS-VQA in multi-round active learning with principled stopping criteria and cost-aware labeling (\eg, mixing absolute category rating, paired comparison, and lightweight screening), enabling scalable dataset growth for emerging regimes such as AI-generated and streaming videos while maintaining annotation reliability and efficiency.

\section*{Acknowledgments}
This work was supported in part by the Hong Kong ITC Innovation and Technology Fund (9440379 and 9440390), and a Google Gift Fund (9220141). We thank Tianhe Wu for assistance with the preparation of Figs.~\ref{fig:teaser} and~\ref{fig:overview}.

{
    \small
    \bibliographystyle{ieeenat_fullname}
    \bibliography{main}
}


\clearpage
\setcounter{page}{1}
\maketitlesupplementary

\section*{Content}
  \vspace{5pt}
This supplementary material provides:
  \vspace{5pt}
\begin{itemize}
  \item Additional implementation details of the experimental pipeline in Sec.~\ref{sec:more details};
  \vspace{5pt}
  \item Additional qualitative gMAD comparisons against competing methods in Sec.~\ref{sec:more gMAD viz};
  \vspace{5pt}
  \item Qualitative visualizations of representative failure samples identified by the failure predictor in Sec.~\ref{sec:failure viz}. 
\end{itemize}

\section{Additional Implementation Details}
\label{sec:more details}

This section supplements Sec.~\ref{sec:implementation details} by describing the detailed training.
We first train VisualQuality-R1~\cite{wu2025visualquality} on the YouTube-UGC training set to obtain a base VQA model. For rapid validation under limited compute, we adopt the default LoRA configuration of VLM-R1~\cite{shen2025vlm}, using group size $K=6$, LoRA rank $r=64$, and a per-GPU batch size of $24$, yielding an effective batch size of $48$ with $2$ gradient accumulation steps. While LoRA can affect the absolute performance of the base model, it does not compromise the comparative validity of our study, because our evaluation focuses on relative improvements across different selection strategies. During training, each video pair is fed to VisualQuality-R1 together with a structured text prompt (see Table~\ref{tab:vqr1 prompt}), producing scalar outputs that are used to compute rewards and optimize the VQA model.

\begin{figure*}[!t]
	\centering
	\includegraphics[width=\textwidth]{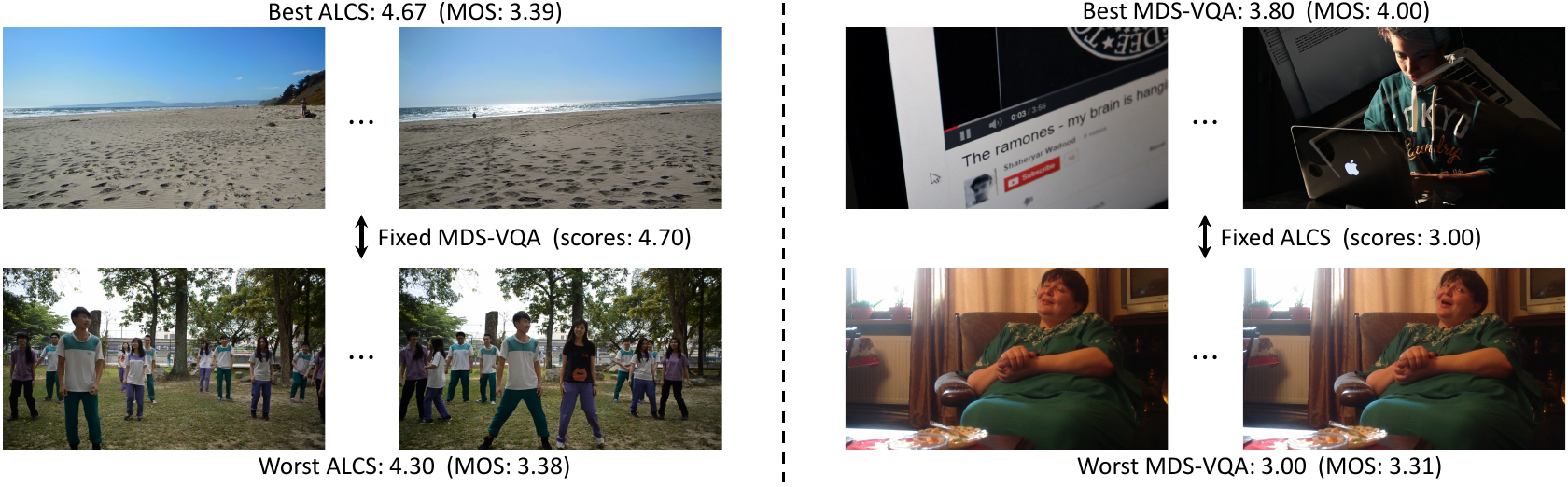}
	\caption{Representative gMAD pairs between VQA models induced by MDS-VQA and ALCS~\cite{yan2022clustering}.
    }
	\label{fig:gMAD supp1}
\end{figure*}

\begin{figure*}[!t]
	\centering
	\includegraphics[width=\textwidth]{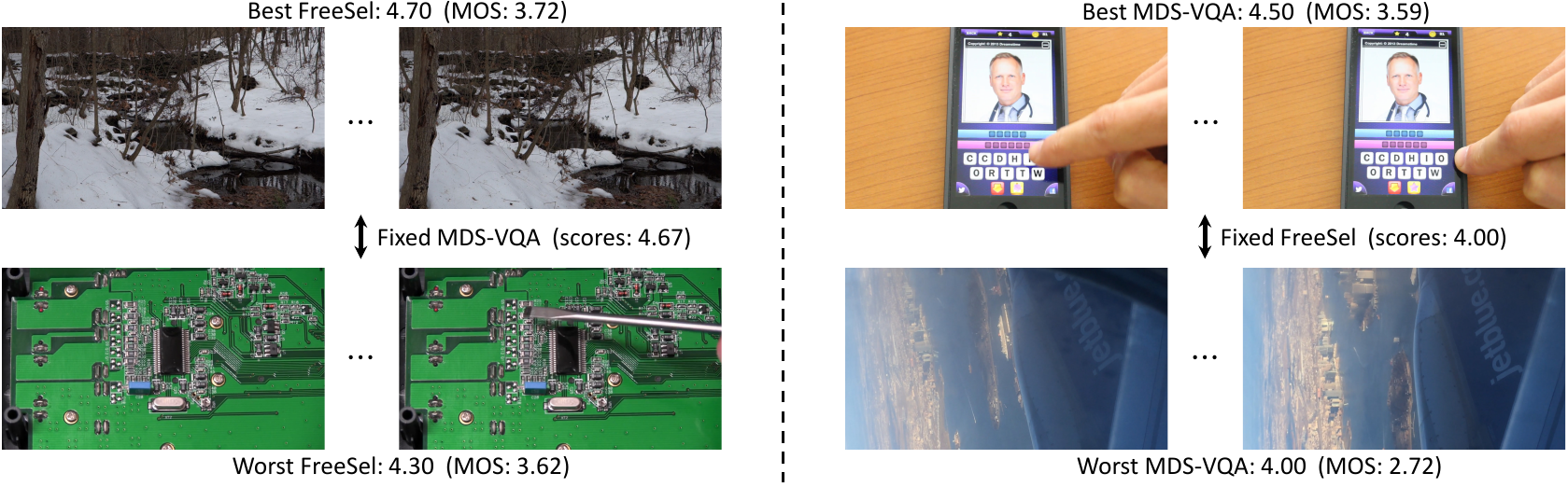}
	\caption{Representative gMAD pairs between VQA models induced by MDS-VQA and FreeSel~\cite{xie2023towards}.
    }
	\label{fig:gMAD supp2}
\end{figure*}

\begin{figure*}[!t]
	\centering
	\includegraphics[width=\textwidth]{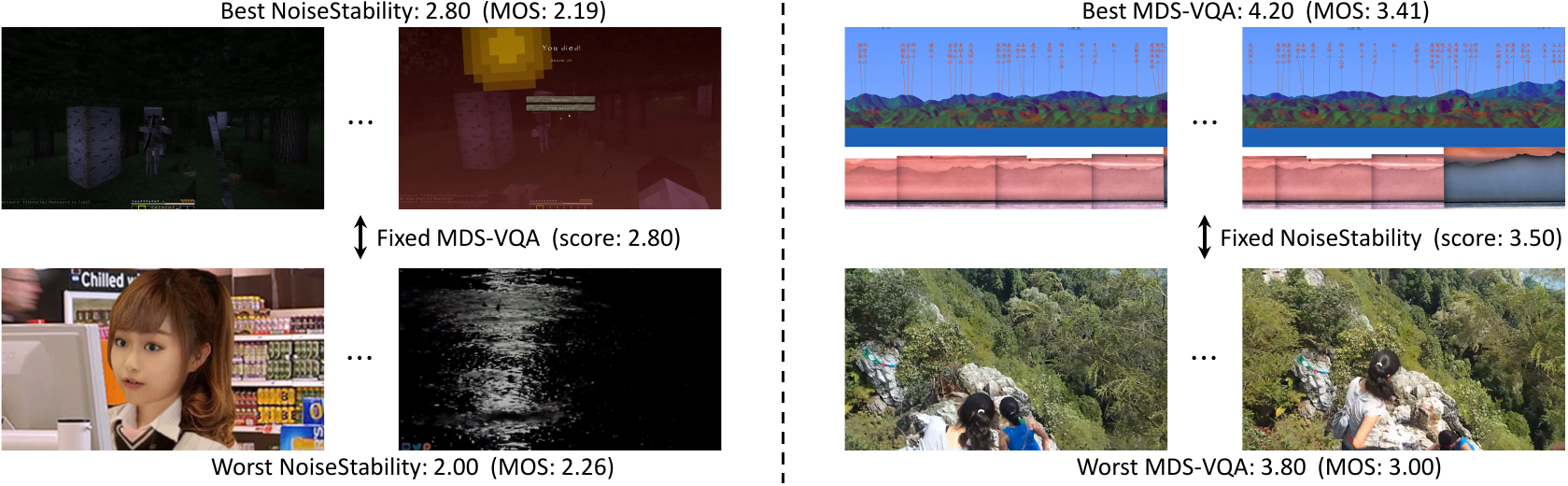}
	\caption{Representative gMAD pairs between VQA models induced by MDS-VQA and NoiseStability~\cite{li2024deep}.
    }
	\label{fig:gMAD supp3}
\end{figure*}

\begin{figure*}[!t]
	\centering
	\includegraphics[width=\textwidth]{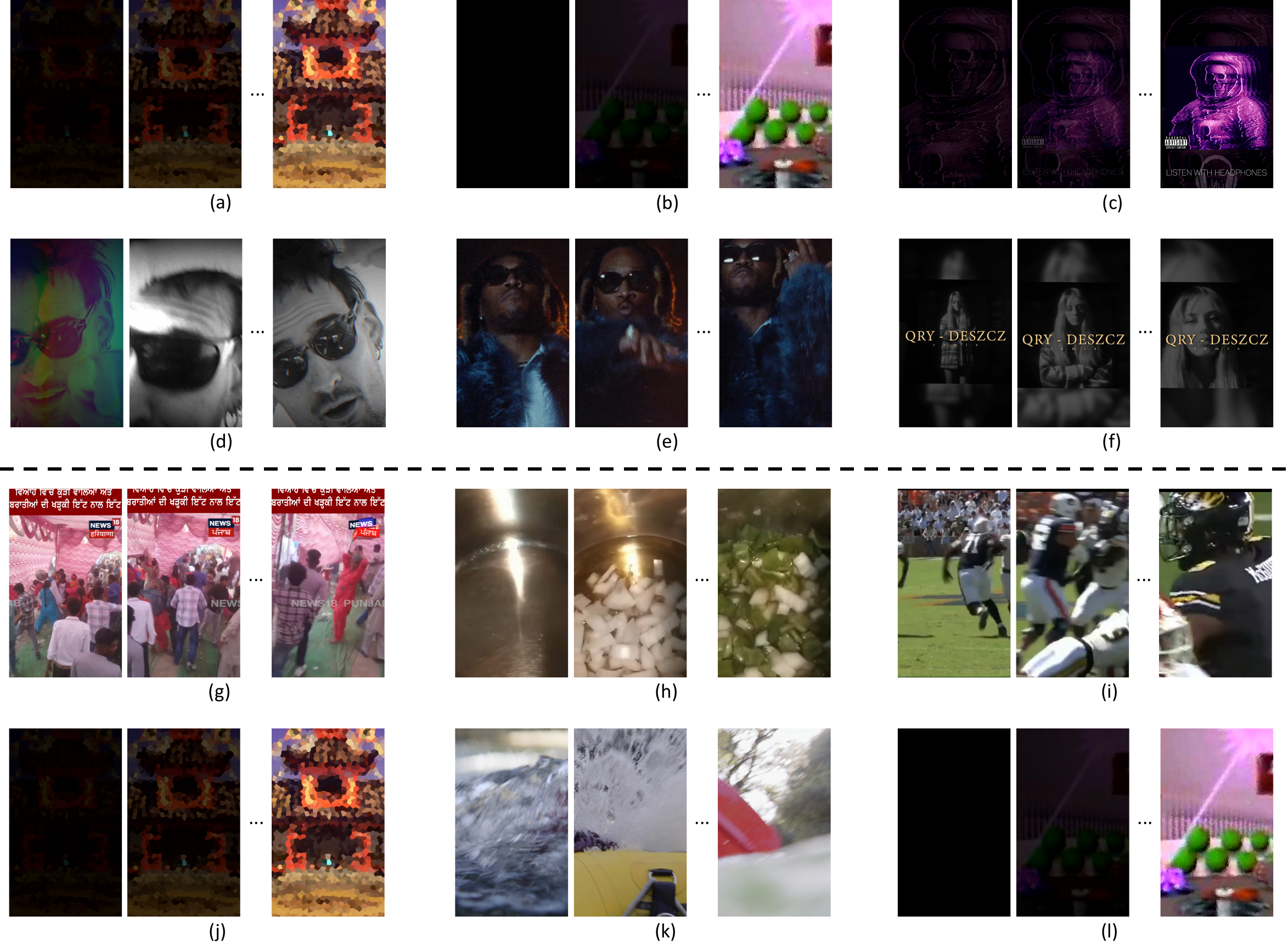}
	\caption{Representative challenging videos from YouTube-SFV SDR~\cite{wang2024youtube} selected by MDS-VQA. (a)-(f) show samples chosen without the diversity term, whereas (g)-(l) show samples chosen with diversity, resulting in a broader coverage of content and distortion patterns.
    }
	\label{fig:failure samples}
\end{figure*}

\section{Additional gMAD Pairs}
\label{sec:more gMAD viz}
Figs.~\ref{fig:gMAD supp1}-\ref{fig:gMAD supp3} provide additional representative gMAD pairs to further compare fine-tuned models induced by MDS-VQA against those by ALCS~\cite{yan2022clustering}, FreeSel~\cite{xie2023towards}, and NoiseStability~\cite{li2024deep}. Across these comparisons, the MDS-VQA-induced model more consistently exposes distinct failure modes of competing methods (when acting as the attacker), and remains more robust under attacks (when acting as the defender), yielding predictions that better agree with human perception of video quality.

\begin{table}[!t]
\centering
\renewcommand{\arraystretch}{1}
\caption{Structured text prompt used for training $f(\cdot)$.}
\label{tab:vqr1 prompt}
    \resizebox{\linewidth}{!}{
    \begin{tabular}{p{\linewidth}}
    \toprule
    You are doing a video quality assessment task. \\
    Here is the question: What is your overall rating on the quality of this video? The rating should be a float between $1$ and $5$, rounded to two decimal places, with $1$ representing very poor quality and $5$ representing excellent quality.\\
    First output the thinking process in \texttt{<think>} \texttt{</think>} tags and then output the final answer with only one score in \texttt{<answer>} \texttt{</answer>} tags. \\
    \bottomrule
    \end{tabular}
}
\end{table}

\section{Visualizations of Failure Samples}
\label{sec:failure viz}
Fig.~\ref{fig:failure samples} visualizes representative challenging videos selected from YouTube-SFV SDR~\cite{wang2024youtube} by the proposed MDS-VQA. Without an explicit diversity constraint, the selection tends to concentrate on visually similar hard samples that share a common failure cause. In Figs.~\ref{fig:failure samples}(a)-(f), this manifests as many black-toned scenes where the base VQA model predicts less reliably. After incorporating the diversity term, the selected set becomes substantially broader in both semantic content and distortion characteristics (see Figs.~\ref{fig:failure samples}(g)-(l)), indicating the combined ``hard-and-diverse'' criterion better covers complementary failure modes under the same labeling budget.

\end{document}